\documentclass[conference]{IEEEtran}
\IEEEoverridecommandlockouts
\usepackage{cite}
\usepackage{amsmath,amssymb,amsfonts}
\usepackage{algorithmic}
\usepackage{graphicx}
\usepackage{textcomp}
\usepackage{xcolor}
\usepackage{hyperref}
\usepackage{url}
\def\BibTeX{{\rm B\kern-.05em{\sc i\kern-.025em b}\kern-.08em
    T\kern-.1667em\lower.7ex\hbox{E}\kern-.125emX}}
\begin{document}

\title{Harnessing Metaverse to Revitalize Surgical Practices: Opportunities, Challenges, and Lessons Learned }
\title{SurgMeta: Challenges and Opportunities in Leveraging Metaverse to Revitalize Surgical Practices}
\title{Integrating artificial intelligence and mixed reality in interventional care: A case study of surgical metaverse }
\title{Revitalizing Interventional Care using Immersive Reality and Artificial Intelligence: The Need for Robust Surgical Metaverse}
\title{Towards Realization of Secure and Robust Surgical Metaverse: Opportunities and Challenges}
\title{Revitalizing Interventional Care: The Need for Secure and Robust Surgical Metaverse}
\title{Revitalizing Interventional Care Through\\ Secure and Robust Surgical Metaverses}
\title{Harnessing Secure and Robust AI-XR Surgical Metaverses to Revitalize Interventional Care}
\title{Secure and Robust AI-XR Surgical Metaverses to Revitalize Interventional Healthare}
\title{Revitalizing Interventional Healthcare with \\Secure and Robust AI-XR Surgical Metaverses}
\title{Can We Revitalize Interventional Healthcare with AI-XR Surgical Metaverses?}


\author{Adnan Qayyum$^{1,3}$, Muhammad Bilal$^{2,*}$, Muhammad Hadi$^3$, Paweł Capik$^2$, Massimo Caputo$^4$, Hunaid Vohra$^4$, \\ Ala Al-Fuqaha$^5$, and Junaid Qadir$^{6,*}$ \thanks{$^*$ Corresponding authors.} \\
$^1$University of Glasgow, Glasgow, United Kingdom \\
$^2$University of the West of England, Bristol, England \\
$^3$Information Technology University, Lahore, Pakistan \\
$^4$Bristol Heart Institute, University of Bristol, Bristol, England \\
$^5$Hamad Bin Khalifa University, Doha, Qatar \\
$^6$Qatar University, Doha, Qatar\\ }


\maketitle

\begin{abstract}
Recent advancements in technology, particularly in machine learning (ML), deep learning (DL), and the metaverse, offer great potential for revolutionizing surgical science. The combination of artificial intelligence and extended reality (AI-XR) technologies has the potential to create a surgical metaverse, a virtual environment where surgeries can be planned and performed. This paper aims to provide insight into the various potential applications of an AI-XR surgical metaverse and the challenges that must be addressed to bring its full potential to fruition. It is important for the community to focus on these challenges to fully realize the potential of the AI-XR surgical metaverses. Furthermore, to emphasize the need for secure and robust AI-XR surgical metaverses and to demonstrate the real-world implications of security threats to the AI-XR surgical metaverses, we present a case study in which ``an immersive surgical attack'' on incision point localization is performed in the context of preoperative planning in a surgical metaverse.
\end{abstract}



\section{Introduction}

The metaverse, a more advanced and immersive version of the Internet where virtual and physical worlds are seamlessly integrated, is rapidly gaining momentum. This is being enabled by the combination of artificial intelligence (AI) with immersive extended reality (XR) technologies such as virtual reality (VR), augmented reality (AR), and mixed reality (MR). These XR technologies complement each other: AR projects digital content (text, images, and sounds) onto the real world; VR creates a computer-simulated environment that blocks out the real world through a VR headset; while MR merges the real and physical worlds to create environments and experiences where physical and digital objects coexist and interact in real-time. The AI-XR metaverse is expected to have a significant impact in the field of healthcare, particularly for revitalizing surgical practices through the use of AI-XR technologies for clinical care, medical training, and patient care in future \textit{surgical metaverses}, as highlighted in recent literature \cite{van2022robotic}.


The trend of using immersive technologies in surgical practices (e.g., using AR) is becoming increasingly common. Giannone et al. presented a review of how AR and AI can be used in minimally invasive robotic liver surgery \cite{giannone2021augmented}. Similarly, Tan et al. offer a comparative perspective in their paper focused on analyzing the opportunities and challenges of virtual healthcare and metaverse in ophthalmology in \cite{tan2022metaverse}. The use of VR and MR in surgical practice has also been successfully demonstrated recently, as demonstrated by Dr. Robert Masson, who performed the first reconstructive surgery in an MR-based surgical operating theatre. While there is significant potential to leverage XR and AI in robotic surgery, technical challenges such as lack of tactile feedback still remain that might hinder the smooth development and operation of AI \cite{van2022robotic}.

The true potential of the AI-XR surgical metaverse can be only fully realized when such systems are considered dependable. This is particularly important in interventional care, as any harm caused by technology cannot be reversed due to the invasive nature of the procedures. Surgical errors can have severe consequences, particularly in domains like cardiac surgery, leading to lifelong complications or even patient death. This demands that the safety, security, and robustness of these AI-XR surgical metaverses should be fool-proof. 

In this study, we aim to offer a vision of a future secure and robust surgical metaverse and provide insightful perspectives on potential applications and challenges. The major contributions of this paper are as follows:

\begin{enumerate}
\item We investigate the applications of the AI-XR surgical metaverse that can revitalize interventional healthcare.
\item We delve into the obstacles that could impede the long-term development of the AI-XR surgical metaverse.
\item We provide a case study exemplifying a real-world instance of an attack named ``an immersive surgical attack'' in a virtual preoperative scenario to emphasize the need for the development of a secure, safe, and robust AI-XR surgical metaverse.
\end{enumerate}

\section{Background}
\label{sec:back}

\subsection{Metaverse: An Introduction}
The term ``metaverse'' itself is the composition of two terms: (1) `meta', which means transcending; and (2) `universe', which in this case refers to the physical universe. The metaverse is an emerging technology that combines the actual world, virtual reality (VR), and augmented reality (AR) to provide a fully immersive, seamless experience with the real and virtual worlds simultaneously. While there is no universally agreed-upon definition for the term metaverse in the literature \cite{cheng2022will}, in general, the metaverse envisions a virtual extension of the physical universe in which human beings can seamlessly interact with surrounding objects in an immersive way to unlock unlimited possibilities. Metaverse technology will enable the creation and interconnection of multiple virtual universes over the Internet, where the users can have immersive social experiences similar to the physical world. The metaverse will be based on several enabling technologies including the Internet, AI, XR (in its various forms, VR, AR, MR), blockchain, machine learning (ML), deep learning (DL), Web3, and 6G communication and sensing devices such as wearables, to mention a few. We refer interested readers to a comprehensive survey focused on a detailed discussion of the metaverse and related concepts \cite{lee2021all}.  

\subsection{AI in Surgical Metaverse and Surgical Data Science}

AI is essential for surgical metaverses, with potential applications including natural language processing, computer vision, network communication, blockchain, and digital twins \cite{qayyum2022secure}. Surgical data science, using advanced AI techniques, has emerged as a discipline in recent years \cite{maier2022surgical}. AI-assisted surgical practice not only provides surgeons with critical information, but also enables promising capabilities such as surgical skills assessment, minimally invasive surgery, and AI-guided surgeries using VR and AR. Moreover, personalized surgical interventions, which are challenging to identify manually, can be modeled using ML/DL. The AI-XR surgical metaverse can revolutionize interventional care by using AI to improve the performance of all clinical stakeholders involved in surgical practice and improve patient outcomes. AI-empowered data analytics is key to personalized decision-making, interoperative assistance, customized interventions, situational awareness during surgery, and post-surgery performance assessment. In addition, it can provide surgical training in interventional care by creating computer-simulated surgical environments using generative models and VR. However, a lack of representative (annotated) datasets to train AI models is a major challenge hindering the success of these methods in interventional care.
Gathering such data can enable various powerful applications of surgical data science, including but not limited to:

\begin{itemize}
    \item \textit{Objects Localization \& Surgical Scene Understanding:} Automatically detecting and tracking objects, such as instruments and tissues, from surgical recordings is essential for unlocking many transformative surgical data science applications. For instance, algorithmic instrument tracking can aid in creating automated metrics for measuring surgical performance and task efficiency, identifying proficient instrument use, and planning logistics for a procedure. Object detection and tracking can also help identify objects that were introduced into a patient's body but not removed before the incision was closed. Implementing these localization and scene understanding features requires the use of AI for image classification, segmentation, object detection, and tracking.

    \item \textit{Visual Tracking:} Visual tracking, which involves following a specific area in the endoscopic surgical scene during procedures, is an important image regression task. For example, in heart surgery, surgeons may need to track critical anatomical structures such as the phrenic nerve to avoid damaging it while accessing the heart regardless of the endoscopic view's zoom or pan. The ML model for visual tracking is crucial for developing various useful and exciting applications such as soft tissue deformation estimation, lesion tracking, AR-based real-time predictions, and robotic visual serving. 

    \item \textit{Preoperative to Intraoperative Registration (PIR):} PIR is a technique that involves identifying key landmarks in 2D intraoperative imagery and aligning them with the 3D cardiac CT/MR data from preoperative planning. This process requires two models: one for segmenting cardiovascular organs in cardiac procedures and another for registering 2D anatomical curves to correspond with 3D curves from preoperative planning. These models can enhance the situational awareness of cardiac surgeons by providing AR overlays to accurately localize critical areas (such as the aortic/mitral valve) for removal and subsequently identify optimal suturing locations for synthetic valve implantation in replacement procedures.
\end{itemize}

\section{AI-XR Surgical Metaverse: Applications}
\label{sec:surgmeta}

\begin{figure}[!t]
    \centerline{\includegraphics[width=0.4\textwidth]{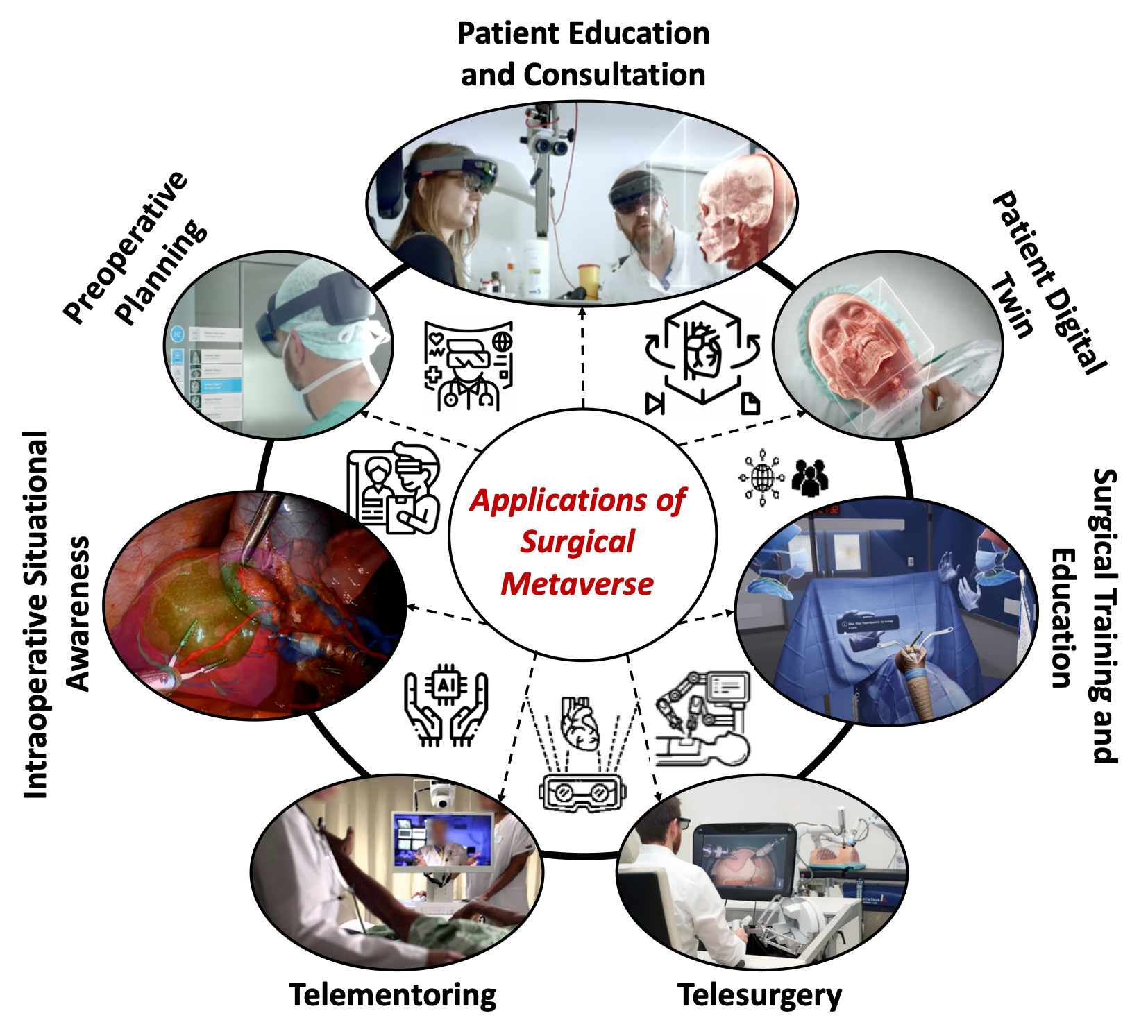}}
    \caption{Illustration of various potential applications that will be featured in the AI-XR surgical metaverse.}
    \label{fig:surgmeta_apps}
\end{figure}

In this section, we examine various promising applications of immersive technologies and AI in current surgical practice (an illustration is presented in  Figure \ref{fig:surgmeta_apps}). 




\subsection{Preoperative Planning}
Preoperative planning refers to the process of analyzing patient data to diagnose patient issues and define the most appropriate interventional strategy for resolution with the least risks and improved patient outcomes. 
The process involves the full surgical team including surgeons, anesthesiologists, nurses, and other operating room (OR) staff. In traditional practice, it is usually done manually in a hand-written fashion where each step is documented for preoperative risk assessment. A well-thought preoperative strategy can help the surgeon perform the procedure more efficiently and with less stress. Generally, preoperative planning consists of three key, interrelated tasks: (1) identification of desired outcome(s); (2) formulating interventional tactic(s); and (3) surgical logistic planning. The emerging immersive technologies empowered with AI can offer promising ways to transform preoperative planning. For example, AI-XR based simulator can be used to visualize and analyze multi-modal patient data for efficient diagnosis and planning procedures and other logistics for the operation. Such a platform can provide the integration of state-of-the-art AI models for interpreting EHRs, MRI, and CT datasets to calculate clinical parameters informing preoperative decisions. Surgeons will glance through the entire patient's medical history, e.g., previous surgery or an injury can also be considered in preoperative planning in a more intuitive visual analytics interface. They will see and modify details about key surgical tasks like incision locations, landmarks for tissue removals, anesthesia or inotrope administration doses, etc. When it comes to surgical logistics planning, such a platform can magnificently help surgeons with describing the end-to-end patient journey, surgical flow, instruments of right lengths, angles and curvatures, personalized risk mitigation strategies, duration of OR, ICU, and ward stays, and types of beds during stay with right provisions. Surgeons can harness text-generation AI to quickly write brief instructions for peers or patients. Last but not least, such a platform can allow surgeons to perform the entire procedure virtually in collaboration with the surgical team to fully understand patient-specific complexities, and rehearse the procedure several times for proficiency before stepping into the OR. 


\subsection{Patient Consultation and Education}
Surgeons evaluate patient data prior to meeting them during the consultation and education stage. This meeting typically involves a detailed discussion of the diagnosis, patient suitability for the procedure, therapeutic benefits, potential risks of the planned surgery (open vs minimally invasive), and addressing patients' questions so that they can make informed decisions and express consent befitting their situation. It may involve informing patients about what to expect during the recovery period or how to get ready for the surgery, such as by following particular dietary or medication instructions. It helps in ensuring that the patient is fully informed and prepared for the surgery which is reported to reduce patient anxiety and improves outcomes of the interventional care \cite{pearce2006documenting}. In the current clinical practice, surgeons conduct multiple one-to-one sessions to educate their patients about alternative interventional options along with briefings about their merits and demerits, including patient-specific risks. The use of immersive technologies can significantly improve patient consultation and education tasks for surgeons. For example, in valve-related procedures, AR can be used to visualize the 3D models of a patient's cardiovascular system where surgeons and patients can interact during the consultation to explain the diagnostics, the origin and size of the leak, quantification of the degree of the regurgitation, and the demonstration of what will be done during the operation through the simulated procedure. Further, patients can see the entire preoperative plans including risk mitigation strategies in 3D-powered AR overlays, personalized to their anatomies and health conditions.

\subsection{Patient Digital Twin}
AI-XR-empowered surgical metaverse will allow for the creation of a digital twin of patients in the virtual world. Where the biomarkers of the patients can be sensed and mapped on the digital twin in real-time using a range of data streaming \& sensing technologies. The most fascinating aspect of the digital twin will be the generation of patient-specific organ systems to model different anatomical variations and features. For example, advancements in generative ML/DL can be leveraged to reconstruct a digital replica of a patient's cardiac system in 3D. In this regard, multi-modal data (including MRI, CT, X-rays, etc.) can be used to efficiently model fine-grained characteristics of the organs in the digital twin. After the successful creation of a digital replica of patient organ systems, the AI-XR surgical metaverse will allow unprecedented opportunities that have never been possible before including counterfactual reasoning (CR), scenario modeling, and treatment optimization, as elaborated next.  

\subsubsection{Counterfactual Reasoning (CR)} CR using a patient digital twin can be a useful tool to evaluate the potential benefits and risks of alternative interventional procedures before embarking on the actual surgery. For example, surgeons may use CR to identify the potential risks and outcomes of planned procedures by simulating it in the AI-XR surgical metaverse on the patient's digital twin. It can be used to answer important questions about a particular interventional task, e.g, what may happen if we do not do it in this way. This can help the surgeons in deciding the optimal interventional procedure before performing surgery in the OR. 

\subsubsection{Scenario Modelling} Surgeons can use scenario modeling to create and analyze several scenarios to improve the planned surgical procedure. The patient's anatomy and the surgical tools and equipment that will be utilized during the operation can be modeled virtually using scenario modeling. It can also be used to determine potential complications and challenges that could occur during the surgery, allowing the surgical team to develop contingency plans to address such issues. For example, a surgeon may use scenario modeling to identify the optimal incision points during dissection for minimally invasive cardiac surgery while considering the anatomical structure of the patient's ribs and internal organs. 

\subsubsection{Post-operative Care} Once surgery has been successfully performed, the digital twin feature in the AI-XR surgical metaverse will allow surgeons to simulate the impact of postoperative care in real time and optimize treatment decisions with greater efficacy. For example, patients are prescribed inotropes to improve hemodynamic function and avoid hypotension and Low Cardiac Output Syndrome (LCOS) after heart surgery. The prescribing surgeons can simulate the impact of various doses of inotropes in the AI-XR surgical metaverse and visualize risks in the digital twin for various adverse reactions to determine the most optimal dosages.

\subsection{Intra-operative Situational Awareness}
The use of ML/DL together with immersive technologies like AR can improve the intraoperative situational awareness of surgeons. Several retrospective studies have demonstrated that computer-guided intraoperative awareness reduces operation time and eliminates certain risks \cite{giannone2021augmented}. Most preoperative plans performed on 3D medical images (CT and MRI scans) can be projected into the surgical scene in real-time by translating landmarks for tissue manipulation over the patient’s actual anatomy, which can revolutionize the care quality and surgeons’ performance. Following AR-guided instructions during the surgical phase is key to avoiding errors that might lead to irreversible complications. For example, image-guided cardiac surgical resection can eliminate the risk of damaging critical anatomical structures like the phrenic nerve in cardiac procedures. Visual tracking can allow surgeons to keep track of such arbitrary anatomical areas in dynamic and ever-evolving surgical scenery. In addition, efficiently developed ML/DL-based models for performing specialized procedural tasks by analyzing surgical footage in real time can significantly revitalize interventional delivery practices. Features like presenting checklists generated from the planned surgical flow and real-time task tracking in AR throughout the surgery can boost surgeon performance and avoid possible human errors, which are associated with working in such intense clinical environments. Using a computer vision-assisted suturing inspection via AR can provide surgeons with instant feedback on the quality of stitches in aortic coarctation and the risk of bleeding after heart surgery. This all can lead to the development of visual analytics features in the AI-XR surgical metaverse that is powered by context-aware predictive analytics to provide timely instructions and actionable insights to avoid potentially irreversible damage to patients during the procedure by constantly analyzing multimodal datasets arising from diverse sensing devices attached to the patient.

\subsection{Surgical Training and Education}

The use of VR and AR technologies is expected to revolutionize medical education, training and procedures \cite{thomason2021metahealth}. However, assessing surgical skill gaps and providing personalized training for both trainee and specialist surgeons remains a challenging task. Existing surgical skills assessment and appraisal tools, such as OSATS, GOALS, and GEARS, measure technical skills using global metrics like tissue handling, instrument manipulation, and depth sensing, which are not specific enough to identify areas for improvement and provide personalized feedback. Additionally, there are currently no digital tools that can measure the performance of all team members involved in a cardiac surgery procedure. Even with successful surgeries, there may be room for improvement for certain individuals.

Most surgical AI models lack generalizability because surgical tasks vary between specialties and procedures. For example, models trained for aortic coarctation would not work for assessing performance in mitral valve repair (MVR) procedures, as the tasks are completely different. Furthermore, there are significant variations in the way different surgeons perform the same procedure. There is a significant gap in creating reliable ML/DL solutions for analyzing long surgical videos to rate surgeon performance and identify skills gaps. Moreover, training surgeons on actual patients or cadavers or 3D-printed cardiovascular objects after a known skill gap is impractical due to ethical and economic reasons.

The AI-XR surgical metaverse has the potential to provide an innovative solution for medical students and early-stage surgeons to receive training in a virtual environment, with varying difficulty levels tailored to their skills and expertise. Allowing them to virtually immerse themselves in the patient's body for better learning outcomes. For example, a complete organ system (e.g., the cardiac system) can be visualized outside of the body using XR, allowing surgeons to thoroughly examine patient organs using a photo-realistic imaging interface. Novice surgeons can train on different interventional procedures virtually, on digital twins of patients, and immediately see the outcomes without harming real patients. The AI-XR surgical metaverse also offers the possibility of infinitely repeatable, personalized training. For example, if a surgeon has difficulty with aortic valve replacement in patients with congenital heart disease, they can practice in a virtual environment to understand and test the anatomical variations of patients, ensuring they are comfortable with the procedure before performing it on real patients. Furthermore, this virtual training environment enables surgeons to practice high-stakes procedures safely and repeatedly, reducing the potential for errors and increasing the overall quality of care.

\subsection{Telementoring}
Telementoring in surgical practice is a type of remote mentorship where an experienced surgeon guides the surgical team remotely during a procedure. It enables pre and intraoperative knowledge sharing and consultation among surgeons, making it a major application in the AI-XR surgical metaverse. It is particularly useful for performing complex surgeries in less developed areas where expertise is scarce and for remote surgical training. Telementoring can be implemented using videoconferencing software or enhanced by the AI-XR surgical metaverse. It offers potential to scale and improve surgical care while improving patient outcomes. However, there are many challenges that can hinder its adoption, including ethical considerations (i.e., if a surgeon is not present in the OR, it can raise serious questions about responsibility and liability). This calls for the establishment of protocols and guidelines that must be followed by medical experts during telementoring. 

\subsection{Telesurgery: AI-XR Guided Robotic Surgery}
The robotic surgical interface provides hybrid functionalities to facilitate pre and intraoperative processes, e.g., image-guided surgical resection. Robotic surgery has been adopted in liver resection (a surgical procedure to remove a part of the liver) that enables minimally invasive procedures. However, the literature shows that despite numerous series, it lacks strong evidence \cite{giannone2021augmented}. The AI-XR surgical metaverse will take robotic surgery to an unprecedented level by allowing expert surgeons to remotely participate in the surgery in a collaborative virtual environment, where the robot will be performing the surgery in real-time in the OR. The literature provides some evidence of telesurgery on animals supporting its feasibility. For example,  Mines et al. \cite{mines2007feasibility} discussed successful telesurgery in the ophthalmological domain, where corneal lacerations repair was performed on rabbit eyes. The authors concluded that the surgeons were able to successfully undergo corneal repair using telesurgery. However, the time of telesurgery was much larger than the manual surgery, i.e., on average 80 and 8 minutes, respectively.

\section{Challenges Hindering AI-XR Surgical Metaverse}
In this section, we elaborate on a number of persisting challenges that could hinder the development of the AI-XR surgical metaverse. 


\subsection{User-Specific Dynamics and Versatile Interaction Support}
One of the key challenges in the AI-XR surgical metaverse will be to model user-specific dynamics and to provide support enabling versatile interactions. In surgical science, it is well-established that the key tasks even in a single surgery may vary according to the patient anatomical variations and current condition. Moreover, surgical tactics can be highly subjective, i.e., individual surgeons might perform a specific surgical task in a variety of ways. Therefore, the development of an automatic feature in the AI-XR surgical metaverse becomes challenging as to how to model such versatile and dynamic interactions to provide predictive analytics for informed decision-making. One natural approach to overcome this issue is to train the ML/DL models using such data that is acquired from the surgeries of a diverse number of surgeons. However, realistically it is not feasible. While on the other hand, built-in capabilities to automatically retrain the underlying models when an unseen interaction is observed can be a good approach to addressing this issue. However, it will raise serious questions about the robustness and trustworthiness of such systems concerning the quality of predictions they will make when fine-tuned using such sparse data.

\subsection{Automatic Immersive Overlays Translation}
The literature provides strong empirical evidence supporting the generation of 3D images from 2D images for different modalities such as MRI and CT scans using generative ML/DL models. However, when it comes to the creation of 3D digital twins of patients' organs using ML/DL models in a fully immersive surgical metaverse, the efficacy of such systems can be questioned. For example, how can we translate patient anatomy into 3D VR overlays that support diverse interactions? Moreover, surgeons observe changes in patient dynamics between the pre-operative planning and during surgery. To get a better perspective, for example in the case of congenital heart surgery, the surgeons estimate the anatomical characteristics of the patient's heart in preoperative planning while utilizing information from multi-modal data (when the heart is full of blood). However, the calculations can vary during the surgery when the blood is removed from the heart. This leads to a serious situation and challenge for the surgeons to amend the estimated parameters to ensure the efficacy of the surgical procedure. 

\subsection{High Computational Requirements} 
Data acquisition and processing in the AI-XR surgical metaverse will be a large-scale endeavor, generating zettabytes of data. Storage and processing of such large-scale data are extremely challenging considering the currently available computational resources. Moreover, the AI-XR surgical metaverse will be empowered with different ML/DL models that will be trained and inferred using such huge datasets thus limiting their real-time predictions. To get a perspective on this problem let us consider the development of an ML/DL-based surgical tool detection system using endoscopic surgical videos of minimally invasive cardiac surgery. Such types of surgeries normally last around 4 hours and endoscopic videos are usually acquired at a frame rate of 60FPS. So, it will roughly generate 60$\times$60$\times$60$\times$4 2D images that can be used for training AI models. Moreover, the AI-XR surgical metaverse will require high bandwidth communication to provide seamless operations be it for mimicking real-time surgery or enabling telesurgery. 

\subsection{Secure and Real-Time Communication}
Ensuring the security and integrity of data collected and processed will be a major challenge in the AI-XR surgical metaverse. Any slight tampering in the data stream can lead to unavoidable, life-threatening consequences. Moreover, data communication in the AI-XR surgical metaverse will have stringent QoS and real-time requirements, as any delay in the delivery of data packets or loss of data packets can potentially lead to serious emergency situations threatening the life of the patient. Therefore, the AI-XR surgical metaverse requires dedicated and customized communication protocols to meet the stringent requirements even in the presence of strong network connections, e.g., 5G and 6G communication.   

\subsection{Availability of High-Fidelity Clean Data}
To develop a robust AI-XR surgical metaverse enabling unprecedented opportunities in interventional care, we need a substantial amount of clean and annotated data. Similar to any AI-empowered automated system such as self-driving cars, the performance of the AI-XR-empowered surgical metaverse will highly depend upon the availability of high-fidelity clean data. The acquisition of large-scale and diverse surgical data can lead to the development of ML/DL-empowered personalized surgical interventions that will not only assist surgeons in making timely and critical decisions but will also enable computer-assisted minimally invasive surgeries. In addition, such systems will improve situational awareness and will enable automatic surgical skills assessment.
Moreover, to ensure the efficacy of surgical interventions and to avoid the risk of death in surgeries, it is crucial to study and understand surgical complications predisposed by different diseases resulting in anatomical variations in the organs, e.g., arteries in the heart. There are currently no publicly accessible datasets for the scientific community that capture patients’ variability, multi-modal medical data (including medical images, ECG, radiological reports, EHRs, etc.), surgical details, and interventional outcomes. If such data becomes available, it can be used to develop different applications and tools in the AI-XR surgical metaverse, particularly surgical training tools to train surgeons to improve their skills, knowledge, and situational awareness. 

\subsection{Lack of Domain-Specific Annotations Tools}
It is recognized that the majority of ML/DL techniques work in a supervised learning fashion and their performance highly depends on the quality of annotation used in the training \cite{qayyum2020secure}. Moreover, the annotation of medical data is often more challenging due to a number of factors, including expert availability, high annotation costs, and time. In addition, the lack of domain-specific annotations tools will be a major challenge that could hinder the development of the AI-XR surgical metaverse in the longer term. As it will have various features and applications capturing data from almost every aspect of the human body, it will require customized tools to annotate the acquired data to ensure their robust operation. Usually, people tend to use general-purpose ML libraries and tools for the annotation of surgical videos which work satisfactorily to some extent but their efficacy is limited when domain-specific annotations are required. For example, we may require high-quality labels describing clinical decisions, surgical scenery, task quality, and moment-by-moment risks in finer detail, and varying granularity. 
Moreover, the majority of healthcare professionals are unaware of the tools used for the annotation of data, even when willing to annotate the data. This calls for the development of domain-specific annotation tools to facilitate surgeons and medical experts in performing annotations with minimal effort. MONAI is a domain-specific AI platform that has been designed specifically for segmentation pipelines in radiology and pathology. Recently, the platform has added functionality for labeling endoscopic videos, however, it may present challenges for weakly supervised labeling or implementing active learning due to the requirement for users to adhere to the coding standard of the specific ML library supported by the platform.


\subsection{Lack of Integrated Multi-Modal Datasets}

The development of AI-enabled tools in surgical science is hindered by several factors, including the lack of good-quality annotated datasets and the inability to effectively integrate and analyze multi-modal data such as radiology reports, images, genetic/epigenetic data, and electronic records with the outcomes of surgical interventions. Additionally, much of the information exchanged among healthcare workers during patient interactions in the OR is not currently captured in a format that can be used for decision-making. This lack of integration and data standardization leads to inefficiency, lack of interoperability, and limited progress in the application of ML/DL-based predictive analytics in healthcare. This is a pressing issue, as poor decision-making and skills can elevate the risk of re-operation, readmission, and death. Therefore, there is a need to develop automated tools for integrating, analyzing, and standardizing multi-modal data across different departments and hospitals. This is especially crucial as the use of AI-generated data in futuristic tools like the AI-XR surgical metaverse is expected to increase in the future.

\subsection{Adversarial ML Threat}

Recent ML/DL methods have achieved state-of-the-art performance in complex tasks such as object detection, semantic segmentation, and image regression. Successful use of DL methods in the detection and diagnosis of different diseases including ophthalmology \cite{gulshan2016development}, clinical pathology \cite{bejnordi2017diagnostic}, and dermatology \cite{esteva2017dermatologist} has been reported. In some cases, these models have even surpassed human predictions, such as DL outperforming human experts in pneumonia diagnosis using X-ray images \cite{rajpurkar2017chexnet}. However, these models are vulnerable to adversarial examples \cite{qayyum2020secure,qayyum2020securing}, where an adversary can craft imperceptible examples to evade the model. Adversarial ML attacks have been demonstrated for different medical tasks, such as skin cancer, diabetic retinopathy, phenomena classification in chest radiographs \cite{finlayson2019adversarial}, and semantic segmentation in brain MRI \cite{paschali2018generalizability}. The most severe consequences of adversarial ML attacks have been demonstrated in CT scans to conceal lung cancer and introduce cancer in healthy individuals during DL-based reconstruction of CT images \cite{mirsky2019ct}. To fully realize the potential of the AI-XR surgical metaverse, it is crucial to strengthen the ML and DL components against adversarial attacks and other similar threats.

\subsection{Ethical Challenges}

There are a number of ethical concerns associated with the proliferation of digital systems, including the AI-XR surgical metaverse. The literature highlights that Internet-enabled systems are more prone to adversarial actions \cite{qayyum2020securingcloud}, and the metaverse is expected to be the new generation of the Internet. Therefore, it is crucial to develop ethical and trustworthy human-centric systems, particularly when it comes to healthcare services like the AI-XR surgical metaverse. Below are a few prominent ethical challenges that could hinder the realization of the AI-XR surgical metaverse:

\begin{enumerate}
\item \textit{Privacy Preservation}: Protecting the privacy of users (surgeons and patients) and their data, including patients' medical data and data generated by a surgeon before and after surgery, will be a major challenge in the AI-XR surgical metaverse. The AI-XR surgical metaverse will collect a substantial amount of user data to provide different services such as personalized surgical procedures and patient-specific recommendations. This data must be protected from both internal and external threats to ensure patient safety.

\item \textit{Medical Ethics}: All questions concerning medical ethics and ethical data use should be considered in the development of the AI-XR surgical metaverse to ensure that the technology is used to the maximum benefit of patients and society.


\end{enumerate}


\subsection{Regulatory Challenges}
Surgical science has witnessed the disruption of technological innovations, which is evident in the use of robotic surgeries in liver resection. However, the literature highlights that the European medical device regulations (MDR) guidelines make it more challenging to introduce new hardware and software into the OR \cite{van2022robotic}. We anticipate that due to the complex nature of the AI-XR surgical metaverse, its regulation will be even more challenging. We envision that once such an AI-XR surgical metaverse has been successfully developed and proven secure, robust, safe, and trustworthy, its translation into clinical practice will become easier. Moreover, we anticipate that there is a long way to realizing the true potential of a full-fledged AI-XR surgical metaverse in a realistic environment. However, features that do not impact the patients directly, e.g., surgical training in the surgical metaverse can be deployed easily.  


\section{Secure \& Robust Surgical-Metaverses' Importance: A Case Study}
\label{sec:case}

\begin{figure*}[!ht]
    \centerline{\includegraphics[width=\textwidth]{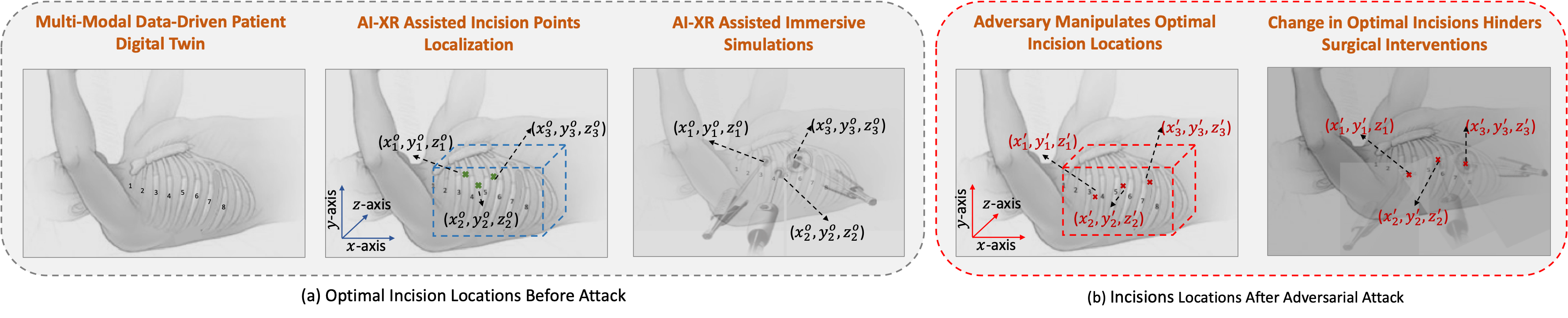}}
    \caption{Illustration of an attack on incision points localization for minimally invasive mitral valve repair (MIMVr) surgery. }
    \label{fig:attack}
\end{figure*}

\subsection{Problem Formulation and Threat Model}


To demonstrate the potential of the AI-XR surgical metaverse, we chose to focus on a case study of preoperative planning for minimally invasive mitral valve repair (MIMVr) surgery. 
In MIMVr, the primary crucial decision surgeons have to make after conducting a risk assessment is to determine the optimal location for small incision points. To ensure precision, efficiency, and safety in MIMVr surgery, surgeons must determine the optimal location for small incision points and have the ability to access and manipulate relevant tissues without damaging surrounding structures. The inability to determine the right incision locations in MIMVr could hinder surgical tasks due to partial visibility of the site or injure critical anatomy or sometimes severe complications leading to conversion to open heart surgery which, for a number of reasons, is undesirable. Using AI-driven volumetric analysis, AI-XR surgical metaverse-based preoperative platform can identify the most optimal locations and size of incisions using patients' MRI and CT scans. Surgeons can explore the prescribed incisions and simulate instrument usage and visibility to carry out the procedure (also, they can override and modify predictions if required). Such planning is then used to create reports to efficiently perform these tasks during the actual surgery. It is important that these decisions are protected as any alterations could jeopardize the efficacy of the  procedure. 
The following attack scenarios are possible.


\begin{enumerate}
    
\item \textit{Adversary maliciously alters the stored optimal incision points.} The first attack, manipulating stored incision points on the surgical metaverse, is quite straightforward. To do so, an adversary can evade the authentication system and insert or replace malicious values stored against a particular patient. This can lead to errors in surgical procedures and can compromise patient safety.

\item \textit{Adversary manipulates the immersive platform.} The second type of attack, manipulating the immersive platform, is more subtle as it involves making imperceptible changes in the virtual environment that lead to the alteration of optimal incision points without being noticed by the human expert. This could be achieved by altering the virtual anatomy, modifying the virtual surgical instruments, or manipulating the virtual haptic feedback. Such an attack can lead to errors in surgical procedures and can compromise patient safety.

\end{enumerate}

\subsection{Attacking Incision Points Localization}
We assume the locations of the optimal incisions required for MIMVr surgery are between the ribs 3-4, 4-5, and 5-6 having cartesian coordinates of $(x_1^o,y_1^o,z_1^o)$, $(x_2^o,y_2^o,z_2^o)$, and $(x_3^o,y_3^o,z_3^o)$, respectively. The immersive simulation capability in AI-XR surgical metaverse-empowered preoperative planning will allow surgeons to validate the efficacy of incision points by simulating the interventional procedure. Specifically, they can insert different tools into the patient's digital twin and can simulate tools usage and other tasks (as shown in Figure \ref{fig:attack} (a)). After performing validation, the user will store these locations and will generate a report to be used in practical surgery. These features are both fascinating and groundbreaking without any adversarial actions. However, it is very trivial to perform an immersive attack on such a system. 

We developed an imperceptible attack in which we introduce minute movements of the patient's digital twin when the human user is selecting optimal incision points. Incision points should be as accurate as possible to ensure smooth surgical tool usage and positive surgical outcome with reduced risks during the surgery. On the other hand, any minute change in these points can cause severe and irreversible consequences. The impact of such an attack is illustrated in Figure \ref{fig:attack} (b), which shows that after an adversarial attack, the incision locations are shifted from ribs 3-4, 4-5, $\&$ 5-6 to 3-4, 5-6, $\&$ 7-8 having different coordinates, i.e., $(x_1^{'},y_1^{'},z_1^{'})$, $(x_2^{'},y_2^{'},z_2^{'})$, and $(x_3^{'},y_3^{'},z_3^{'})$. This makes it difficult to undergo the required surgical procedure effectively, as with the manipulated incisions the subject organ (heart in our case) can not be effectively reached. Also, it makes tool utilization quite difficult, as other internal organs can hinder their movement and can obstruct the endoscopic view. Moreover, the manipulation of optimal incisions can cause significant trouble if remain unnoticed and can lead to open heart surgery. 

We used Unity 3D with C\# programming language to create a surgical environment that shows the operating room, the thoracic cavity of the patient, and surgical instruments (all assets were purchased from Unity assets). The 3D models were created using 3DMax, Maya, and blender software. To experience the surgical metaverse and proposed attack, we used an Oculus Quest 2 VR headset. A demo of our preliminary work can be seen at \href{https://drive.google.com/drive/folders/1l5dXIvU_MS9ingXJn6sFSN-5gL5KMMZc?usp=share_link}{``Immersive Surgical Attack Demo''}, where we demonstrate the impact of manipulating immersive platform on the incision locations in a preoperative planning scenario.  



\section{Conclusions}
\label{sec:concs}

In this paper, we have argued that the incorporation of artificial intelligence (AI) and extended reality (XR) in surgical metaverses has the potential to revolutionize interventional healthcare. We have presented various promising applications of AI-XR surgical metaverses, as well as the challenges that may impede its development. However, in a mission-critical field such as surgical healthcare, where mistakes can have dire consequences, the security and reliability of these metaverses are of paramount importance. Our work has shown that the ``immersive surgical attack'' is possible in a virtual surgical environment, highlighting the need for robust security measures. While the obstacles to fully realizing the potential of AI-XR surgical metaverses are significant, the benefits make it a worthy endeavor. It is essential that the implementation of these technologies prioritizes patient safety and that the benefits outweigh the risks. Our future research will focus on enhancing the functionality of AI-XR surgical metaverses for preoperative planning and making them more resilient against diverse adversarial attacks in different clinical settings, to further understand their capabilities and limitations.

\bibliographystyle{IEEEtran}

\end{document}